\pdfoutput=1
\documentclass{pbmlarxiv}
\usepackage{xcolor}
\definecolor{g}{rgb}{0.0, 0.5, 0.0}
\definecolor{b}{rgb}{0.0, 0.0, 1.0}

\usepackage[normalem]{ulem}
\usepackage{multirow}
\usepackage{minibox}
\usepackage{framed}
\usepackage{makecell}

\newenvironment{framedexample}{\begin{center}\begin{minipage}[c]{1\hsize}\setlength{\FrameSep}{5pt}\begin{framed}\setlength{\parskip}{4pt}}{\end{framed}\end{minipage}\end{center}}
\newcommand{\englishexamples}[1]{\begin{center}\small English translation #1\end{center}}
\usepackage{booktabs}

\begin{document}

% Document title and authors
% ==========================
% Due to journal organization, the article title and authors must be specified
% AFTER \begin{document}. Note that \subtitle is not allowed anymore due to
% problems with indexing in science databases. If needed use colon in the title.

%\title{Analyzing BERT for Diacritics Restoration}
% \title{Diacritics Restoration with BERT}
\title{Diacritics Restoration using BERT with Analysis on Czech language}

% Now put the affiliated institutes first, then author names and the labels
% of their institutes in the "institute" field. The order of institutes and
% authors printed below the title will be the same as the order of your commands.
% Each author can be associated with more than one institute.
% One author must be chosen as the "corresponding author" (using attribute
% corresponding) and his or her email and full address must be provided.
% PLEASE, use Unicode (utf8) encoding (e.g ï instead of \"{i}).
% PLEASE, include country name in the address and
% follow the names listed in https://en.wikipedia.org/wiki/ISO_3166-1

\institute{label1}{Institute of Formal and Applied Linguistics
Charles University, Czech Republic
Faculty of Mathematics and Physics
}

\author{
  firstname=Jakub,
  surname=Náplava,
  institute=label1,
  corresponding=yes,
  email={naplava@ufal.mff.cuni.cz},
  address={Malostranské náměstí 25\\118 00 Praha\\Czech Republic}
}
\author{
  firstname=Milan,
  surname=Straka,
  institute=label1,
}

\author{
  firstname=Jana,
  surname=Straková,
  institute=label1,
}

% If all authors belong to the same institute, you can use simpler syntax:
% \institute{}{Charles University, Faculty of Mathematics and Physics, Institute of Formal and Applied Linguistics}
% \author{firstname=Humpty, surname=Dumpty}
% \author{firstname=Mock, surname=Turtle,
%   corresponding=yes,
%   email={turtle@seacoast.wl},
%   address={Institute of Formal and Applied Linguistics\\
%            Faculty of Mathematics and Physics,\\
%            Charles University\\
%            Malostranské náměstí 25\\
%            118 00 Praha 1, Czech Republic}}
% \author{firstname=Cheshire, surname=Cat}

% The title and authors' names are used in the running head. If they are
% long, you should define short versions. These definitions are optional. You
% define them only if they are needed. The example follows:
\shorttitle{Diacritics Restoration}
\shortauthor{J. Náplava, M. Straka, J. Straková}

% Now print the title by:
\PBMLmaketitle

% Abstract
% ========
% The abstract is placed within the "abstract" environment. It is a mandatory
% part of the article. PLEASE, do not use your own macros in abstract, if possible.

\begin{abstract}
    We propose a new architecture for diacritics restoration based on contextualized embeddings, namely BERT, and we evaluate it on 12 languages with diacritics. Furthermore, we conduct a detailed error analysis on Czech, a morphologically rich language with a high level of diacritization. Notably, we manually annotate all mispredictions, showing that roughly 44\% of them are actually not errors, but either plausible variants (19\%), or the system corrections of erroneous data (25\%). Finally, we categorize the real errors in detail. We release the code at \url{https://github.com/ufal/bert-diacritics-restoration}.
\end{abstract}

\section{Introduction}

Diacritics Restoration, also known as Diacritics Generation or Accent Restoration, is a task of correctly restoring diacritics in a text without any diacritics. Its main difficulty stems from ambiguity where context needs to be taken into account to select the most appropriate word variant, because diacritization removal creates new groups of homonymy. 

Current state-of-the-art algorithms for diacritics restoration are mostly based on either recurrent neural networks combined with an external language model~\cite{naplava2018diacritics, alkhamissi2020deep} or Transformer~\cite{mubarak2019highly}. Recently, BERT~\cite{devlin2018bert} was shown to outperform many models on many tasks while being much faster due to the fact that it uses simple parallelizable classification head instead of a slow auto-regressive approach.

In this work, we first describe a model for diacritics restoration based on BERT and evaluate it on multilingual dataset comprising of 12 languages \cite{naplava2018diacritics}. We show that the proposed model outperforms the previous state-of-the-art system \cite{naplava2018diacritics}
%, which combines bidirection recurrent-neural network with an external language model,
in 9 languages significantly. % and reaches comparative results in the remaining 3 languages.

We further provide an extensive analysis of our model performance in Czech, a language with rich morphology and a high level of diacritization. In addition to clean data from Wikipedia \cite{naplava2018diacritics}, the model was evaluated on data collected from other domains, including noisy data, and we show that stable performance holds even if the text contains spelling and other grammatical errors. 

Sometimes, multiple plausible diacritization variants are possible, while only one gold reference exists, which comes from the original text before diacritization was automatically stripped to create test data. To assess the extent of these cases, we employed annotators to manually annotate all mispredictions and we found that 19\% of errors are plausible variants and 25\% of errors are system corrections of errors in data.

Finally, we further analyse the remaining errors by analysing characteristics of plausible variants.

\section{Related Work}

Diacritics Restoration is an active area of research in many languages: Vietnamese \cite{nga2019deep}, Romanian~\cite{nuctu2019deep}, Czech~\cite{naplava2018diacritics}, Turkish~\cite{adali2014vowel}, Arabic~\cite{madhfar2020effective, alkhamissi2020deep} and many others.

%\footnote{For Arabic diacritization, models need to be slightly modified to allow for character for character swaps.}

There are three main architectures currently used in diacritics restoration: convolutional neural networks~\cite{alqahtani2019efficient}, recurrent neural networks often combined with an external language model~\cite{belinkov2015arabic, naplava2018diacritics, alkhamissi2020deep} and Transformer-based models~\cite{orife2018attentive, mubarak2019highly}. The convolutional neural networks are fast to train and also to infer. However, compared to the recurrent and Transformer-based architectures, they do generally achieve slightly worse results due to the fact that they model long-range dependencies worse. On the other hand, recurrent- and Transformer-based architectures are much slower. 

Recently, the BERT model~\cite{devlin2018bert} comprising of self-attention layers, was proposed and shown to reach remarkable results on a variety of tasks. As it uses no recurrent layers, its inference time is much shorter. 
%Another great benefit, although with no clear relevance to diacritics, is that BERT comes with pretrained weights.
We expect BERT to significantly improve the performance over current state-of-the-art diacritization architectures.

\section{Model Architecture}

\begin{figure*}
    \centering
    \includegraphics[width=.5\hsize]{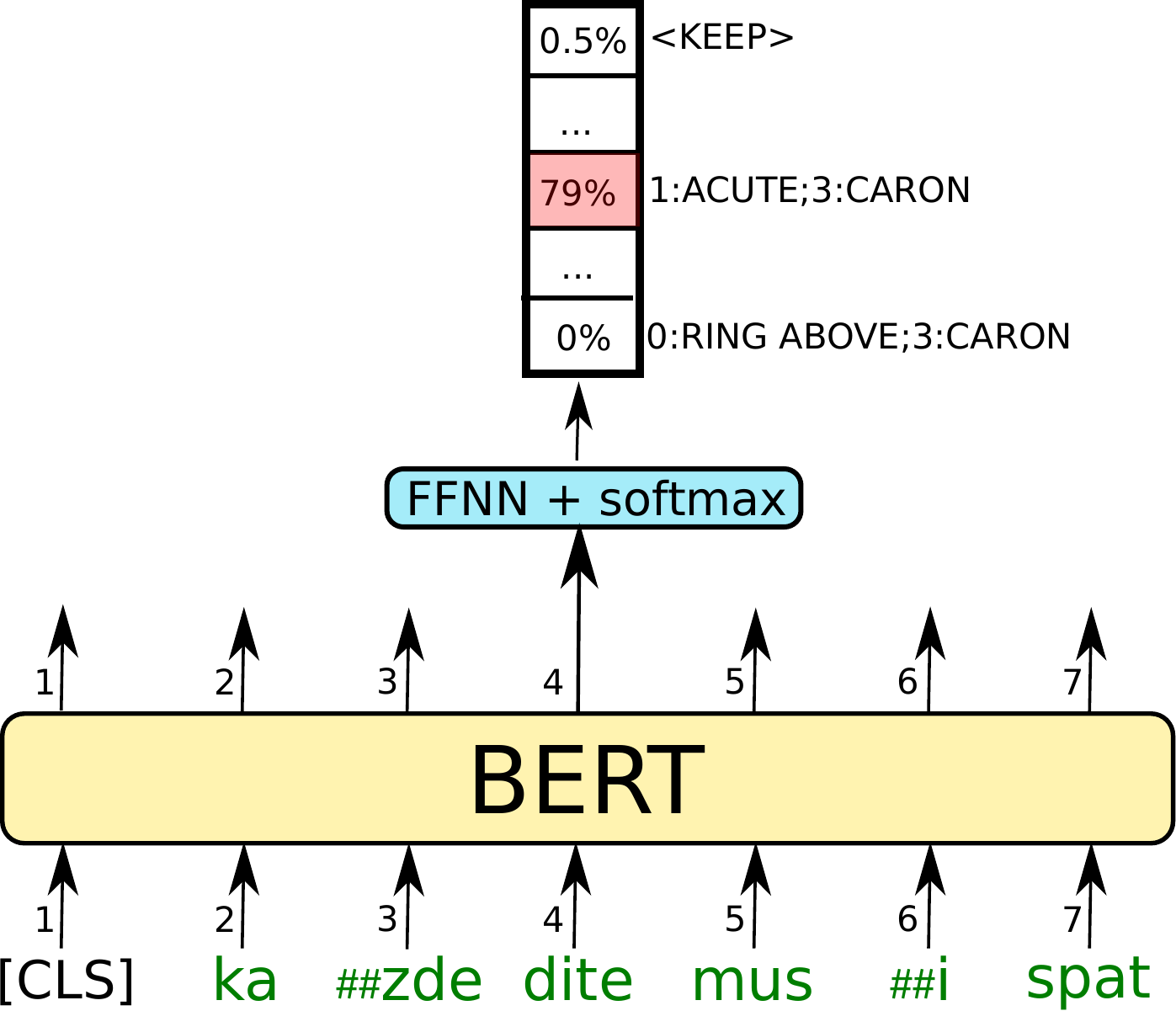}    
    \caption{Model architecture. Text without diacritics, tokenized into subwords, is fed to BERT and for each of its outputs, fully-connected network followed by softmax is applied to obtain the most probable instruction for diacritization. \#\#-prefixes of some subwords are added by the BERT tokenizer.}
    \label{figure_bert}
\end{figure*}

The core of our system is a pre-trained multilingual BERT model that uses self-attention layers to create contextualized embeddings for tokenized text without diacritics. The contextual embeddings are fed into a fully-connected feed-forward neural network followed by a softmax layer. This outputs a vector with a distribution over a set of instructions that define diacritization operation over individual characters of each input token. We select the instruction with maximum probability. The model is illustrated in Figure~\ref{figure_bert}.

\subsection{Diacritization Instruction Set}

To decrease the size of the final softmax layer, the output labels are not the diacritized variants of input subwords, as one would expect, but they are a set of instructions that provide prescription on how to restore diacritics. Specifically, one such instruction consists of index-diacritical mark tuples that define on what index of input subword a particular diacritical mark should be added.

An example of a diacritization instructions set can be seen in Figure~\ref{figure_instructions}. Given an input subword \textit{dite (dítě)}, with four characters indexed from $0$ to $3$, the appropriate diacritization instruction is \textit{1:ACUTE;3:CARON}, in which acute is to be added to \textit{i} and caron is to be added to \textit{e} resulting in a properly diacritized word \textit{dítě}. Obviously, the network can choose to leave the (sub)word unchanged, for which a special instruction \textit{<KEEP>} is reserved. Should the network accidentally select an impossible instruction, no operation is carried out and the input (sub)word is also left unchanged.

To construct the set of possible diacritization instructions, we tokenize the un-diacritized text of the particular training set and align each input token to the corresponding token in the diacritized text variant. The diacritical mark in each instruction is obtained from the Unicode name of the diacritized character. We keep only those instructions that occurred at least twice in a training set to filter out extremely rare instructions that originate for example from foreign words or bad spelling.

%where the whole part after first occurrence of word \textit{WITH} is taken.\js{Poslední větu vůbec nechápu.}

% \begin{table}
% \centering
%     \begin{tabular}{l|l|l}
% Czech & Vietnamese & Latvian \\
% <KEEP> & <KEEP> & <KEEP> \\
% 1:ACUTE &  1:ACUTE & 1:ACUTE \\
% 1:CARON &  2:ACUTE & 2:ACUTE \\
% 2:ACUTE &  1:CARON & 1:CARON \\
% 0:CARON &  1:GRAVE & 0:CARON \\
% 3:ACUTE &  0:CARON & 1:MACRON \\
% 2:CARON &  3:ACUTE & 1:GRAVE \\
% 0:ACUTE &  0:ACUTE & 3:ACUTE \\
% 4:ACUTE &  2:CARON & 2:CARON \\
% 0:CARON;1:ACUTE &  1:DOT BELOW & 0:ACUTE 
% \end{tabular}
%     \caption{Top-10 most occurring instructions for Czech, Vietnamese and Latvian.}
%     \label{table:top_instructions}
% \end{table}

%\begin{figure*}
%    \centering
%    \includegraphics[width=200px]{figures/diakr_instructinos.pdf}    
%    \caption{Diacritization instructions examples for input "dite (dítě)" with 4 %characters, indexed from $0$ to $3$. Index-Instruction tuples are used to generate %diacritics for given subword. Impossible instructions are ignored, even if selected by %the network.}
%    \label{figure_instructions}
%\end{figure*}

\begin{figure}
\centering
    \begin{tabular}{llll}
        input & instruction & result & note \\
        \midrule
        
         d{\color{b}i}t{\color{g}e} & {\color{b}1}:CARON;{\color{g}3}:ACUTE &       d{\color{b}í}t{\color{g}ě} & optimal instruction \\
         
         d{\color{b}i}te & {\color{b}1}:CARON & d{\color{b}í}te & \\
         dit{\color{g}e} & {\color{g}3}:ACUTE & dit{\color{g}ě} & \\
         dite & <KEEP> & dite & no change \\
         di{\color{red}t}e & {\color{red}2}:RING ABOVE & dite & impossible instruction ignored
    \end{tabular}
    \caption{Diacritization instructions examples for input "dite (dítě)" with 4 characters, indexed from 0 to 3. Index-Instruction tuples generate diacritics for given input.}
    \label{figure_instructions}
\end{figure}

\subsection{Training Details}
\label{ssec:training_details}

We train both the fully-connected network and BERT with AdamW optimizer which minimizes the negative log-likelihood. The learning rate linearly increases from 0 to 5e-5 over the first 10000 steps and then remains the same.  We use HuggingFace implementation of \textit{BertForTokenClassification} and initialize \textit{BERT-base} values from \textit{bert-base-multilingual-uncased} model.

We use the batch size of 2048 sentences and clip each training sentence on 128 tokens. We train each model for circa 14 days on Nvidia P5000 GPU and select the best checkpoint according to development set.

\section{Automatic Evaluation on Diacritization Corpus with 12 Languages}

We evaluate our approach on the dataset of \citet{naplava2018diacritics}. This dataset contains training and evaluation data for 12 languages: Vietnamese, Romanian, Latvian, Czech, Polish, Slovak, Irish, Hungarian, French, Turkish, Spanish and Croatian.

We evaluate the model performance using a standard metric, the \textit{alpha-word accuracy}. This metric omits words composed of non-alphabetical characters (e.g., punctuation).

For each language, we compute an independent set of operations and train a separate model. We use the concatenation of the Wiki and the Web training data of \cite{naplava2018diacritics} both for computing a set of instructions and also as the training data for our model.\footnote{In Romanian Web data, ş (LATIN SMALL LETTER S WITH CEDILLA) is for historical reasons often used instead of ș (LATIN SMALL LETTER S WITH COMMA BELOW) and similarly ţ (LATIN SMALL LETTER T WITH CEDILLA) is often used instead of ț (LATIN SMALL LETTER T WITH COMMA BELOW). We replace the occurrences of the previously-used characters (the former ones) with their standard versions (the latter ones).} The size of each instruction set and our results in comparison with the previous state-of-the-art-results of \citet{naplava2018diacritics} are presented in Table~\ref{table:inter_language_results}.
Apart for alpha-word accuracy itself, we also report 95\% confidential intervals computed using bootstrap resampling method.

On 9 of 12 languages, our approach significantly outperforms previous state-of-the-art combined recurrent neural networks with an external language model. The most significant improvements are achieved on Vietnamese and Latvian.

\begin{table}[t]
    \centering
    %\begin{tabular}{l|c|c|c|r}
    \begin{tabular}{l cccr}\toprule
    \multirow{2}{*}{Language} & Instruction & \multirow{2}{*}{\citet{naplava2018diacritics}} & \multirow{2}{*}{Ours} & \multicolumn{1}{c}{Error} \\
    & Set Size & & & Reduction \\\midrule
Czech   &1005&99.06&\textbf{99.22 $\pm$0.046} & 17 \% \\\
Vietnamese  &2018&97.73&\textbf{98.53 $\pm$0.037} & 35 \%	\\\
Latvian &~~720&97.49&\textbf{98.63 $\pm$0.045} & 45 \% \\\
Polish  &1005&99.55&\textbf{99.66 $\pm$0.041} & 24 \%\\\
Slovak  &~~785&99.09&\textbf{99.32 $\pm$0.030} & 25 \%\\\
French  &~~681&\textbf{99.71}&\textbf{99.71 $\pm$0.016} & 0 \%\\\
Irish   &~~189&98.71&\textbf{98.88 $\pm$0.040} & 13 \% \\\
Spanish   &~~492&\textbf{99.65}&99.62 $\pm$0.018 & $-$ 9 \%\\\
Croatian    &~~541&99.67&\textbf{99.73 $\pm$0.018} & 18 \%\\\
Hungarian   &~~767&99.29&\textbf{99.41 $\pm$0.038} & 17 \%\\\
Turkish &1005&\textbf{99.28}&98.95 $\pm$0.046 & $-$ 46 \% \\\
Romanian    &1677&98.37&\textbf{98.64 $\pm$0.056} & 17 \% \\\bottomrule
\end{tabular}
    \caption{Comparison of alpha-word accuracy of our model including 95\% confidential intervals to previous state-of-the-art on 12 languages.}
    \label{table:inter_language_results}
\end{table}

\section{Detailed Analysis on Czech}

We further provide a detailed analysis of our model performance in Czech, a language with rich morphology and a high diacritization level: Of the 26 English alphabet letters, a half of them can have one or two kinds of diacritization marks \cite{NovyEncyklopedickySlovnik}. Czech is also the 4-th most diacritized language of the 12 languages found in the diacritization corpus of \citet{naplava2018diacritics}. 

Particularly, we are interested in the three following questions:

\begin{itemize}
    \item How would our system perform outside the very clean Wiki domain? (Section~\ref{sec:other_domains})
    \item Is it possible that some of the labeled mispredictions are actually plausible variants? (Section~\ref{sec:plausible_variants})
    \item Is there an observable characteristics in the real errors made by the system? (Section~\ref{sec:real_errors})
\end{itemize}

\subsection{Additional Domains}
\label{sec:other_domains}

\begin{table}[t]
    \centering
    \begin{tabular}{lccc}\toprule
    Domain  & Sentences & Words & Evaluated Words \\\midrule
    Natives Formal & 1\,743 & 19\,973 & 19\,138 \\
    Natives Informal & 7\,223 & 99\,352 & 86\,720 \\
    Romi & 1\,490 & 15\,971 & 13\,080 \\
    Second Learners & 5\,117 & 63\,859 & 50\,630 \\\bottomrule
    \end{tabular}
    \caption{Basic statistics of new data for testing diacritics restoration in Czech.}
    \label{table:basic_other_domains}
\end{table}

%\todo{JN: nema v te figure byt boldem jenom vzdy jedna varianta? pokud se spravne divam, tak ta prvni}
%Ano, můžeme tučně vyznačit správnou variantu. Nesprávnou jsem označila hvězdičkou, jak se někdy v lingvistických textech signalizuje.

The testing dataset of \citet{naplava2018diacritics} is composed of clean sentences originating from Wikipedia. It is, however, a well-known fact that the performance of the (deep neural) models may deteriorate substantially when the input domain is changed~\cite{belinkov2017synthetic, rychalska2019models}. To test our system in other, more challenging domains, we used data from a new Czech dataset (unpublished, in annotation process) for grammatical-error-correction that contains data collected from 4 sources:

\begin{itemize}
    \item Natives Formal -- Essays of elementary school Czech pupils (decent Czech proficiency)
    \item Natives Informal -- texts collected from web discussions
    \item Second Learners -- essays of Czech second learners
    \item Romi -- texts of Czech pupils with Romani ethnolect (low Czech proficiency)
\end{itemize}

The dataset covers a wide range of Czech domains. It contains texts annotated in M2 format, a standard annotation format for grammar-error-correction corpora. In this format, each document contains original sentences with potential errors (e.g. spelling, grammatical or errors in diacritics) and a set of annotations describing what operations should be performed in order to fix each error. 

\begin{figure}[t]
    \begin{framedexample}
        Potřebujeme nové idea i \uline{novych} \textbf{lidi}/lidí* , ktery je přinesou . 
        
        Na ulicích vidíme často nekterý lidi , kteří nosí \textbf{barevné}/barevně* \uline{oblečeny}~, které jsou snad hezké , ale určitě nejsou elegantní .
    \end{framedexample}
        
    \itshape
    
    \begin{framedexample}
        \englishexamples{(without ambiguities)}

        We need \uline{new} ideas and also \textbf{people} to come up with them.
        
        In the streets, we can see some people wearing \textbf{colourful} \uline{clothes}, which may be nice but certainly not elegant.
    \end{framedexample}
    \caption{Examples of misleading contexts in noisy texts. Correct diacritization (bold) can only be achieved by grammar corrections of the surrounding words (underlined).}
    \label{table:misleading_context}
\end{figure}

To create target data for diacritics restoration, we apply all correcting edits that fix errors in diacritics and casing. We leave other errors intact, but do not evaluate on words that contain these errors, because they are not directly relevant to diacritics and in many cases, the errors are so severe that evaluation would be controversial. To rule out such words, we create a binary mask that distinguishes between evaluated and omitted words. Although the severely perturbed words are omitted from evaluation, they still remain in the sentence context and may still confuse the diacritization system, making the task potentially more difficult. See examples of such misleading sentence contexts in Figure~\ref{table:misleading_context}.

The basic statistics of the new dataset are presented in Table~\ref{table:basic_other_domains}. We display the number of sentences, the number of all words and the number of evaluated (unmasked) words. Compared to the Wikipedia dataset \cite{naplava2018diacritics}, our new dataset has half the number of sentences and one third of its number of words.

We evaluate our model on all the above introduced Czech domains and present the results in Table~\ref{table:other_domain_results}. Despite our initial concern that the model would perform worse on these domains due to the noisy nature of the data, the results show that the model performance remains roughly stable on all domains. We suppose that although the writers produced quite noisy texts, they at the same time avoided foreign words that are generally harder to correctly diacritize.

\subsection{Error Annotation}
\label{sec:plausible_variants}

\begin{figure}[t]
    \begin{framedexample}
        Nebo záměna kapitol a jejich časová posloupnost v knize je pak ve filmu \textbf{podána/podaná} rozdílně .
        
        Hraní \textbf{šachu/šachů} , ale především karetních her , kritizoval také Petr Chelčický . 
        
        Jeho matka byla \textbf{přadlena/pradlena} , která ke sklonku života propadla alkoholu .
        
        Hororová hudba slouží především pro dokreslení \textbf{filmů/filmu} .
    \end{framedexample}
        
    \itshape
    \begin{framedexample}
        \englishexamples{}
        The chapters and their chronological order in the book are then \textbf{presented/given} differently in the film.
        
        Playing \textbf{a game of chess/games of chess} , but especially card games was criticized by Petr Chelčický . 
        
        His mother was a \textbf{washerwoman/laundress} who fell into alcoholism towards the end of her life .
        
        Horror music is mainly used to complete \textbf{a movie/movies} .
    \end{framedexample}        

    \caption{Examples of ambiguities, each illustrating two diacritization variants (bold), both valid in a given context.}
    \label{table:ambiguties_examples}
\end{figure}

Clearly, removing diacritics creates new groups of homonymy (\textit{dal/dál}, \textit{krize/kříže}). In most cases, the correct diacritization variant can be inferred by a method which takes the sentence context into consideration. However, there are cases, in which more plausible variants are available, e.g., \textit{šachu/šachů}, \textit{pradlena/přadlena}, \textit{podána/podaná}, as illustrated in Figure~\ref{table:ambiguties_examples}. Furthermore, some variants can only be disambiguated in the context of the whole document, such as in: \textit{K nejvýznamnějším patří zmiňované vily/víly.} (more examples in Figure~\ref{table:sentence_vs_whole_context_examples}), not to mention other examples that can be only disambiguated by real-world knowledge such as in \textit{Povrch satelitu/satelitů Země už zkoumalo několik sond}.

However, all our evaluation data are limited only to a single gold reference for each word without diacritics, given by the fact that the gold reference comes from the original text with diacritics. To explore both phenomena among the mispredictions, we hired annotators to examine: a) whether a word is correctly diacritized given the context of current sentence; and b) whether it is correct given a context of two previous sentences, current sentence and two following sentences (thus ruling out the words with even longer document dependencies).

While the evaluation of the clear Wiki data \cite{naplava2018diacritics} is straightforward, some of our newly introduced noisy data may become controversial to evaluate due to erroneous words. Therefore, such words were also marked by the annotators and subsequently removed from our analysis.

An example of a final annotation item presented to an annotator is illustrated in Figure~\ref{table:annotation_item}.

\begin{figure}[t]
    \footnotesize
    \begin{framedexample}
    \begin{tabular}{l|l}
        %Hash:& 48d7a5 \\
        Předpřechozí věta & Popisujeme sítě , které nepoužívají sdílený přenosový prostředek .\\
        Předchozí věta & Přenosové rychlosti se velmi liší podle typu sítě . \\
        Začátek aktuální věty & Začínají na desítkách kilobitů za sekundu , ale dosahují i \\
        \textbf{Aktuální slovo} & \textbf{rychlosti} \\
        Konec aktuální věty & řádu několik gigabitů za sekundu . \\
        Následující věta & Příkladem takové sítě může být internet .\\
        Věta po následující větě & Mezi rozlehlé sítě patří : \\
        Je správně vůči aktuální větě: & Ano \\
        Je správně vůči cel. kontextu: & Ne \\
        Obsahuje překlep: & Ne 
    \end{tabular}
    \end{framedexample}

\itshape
    \begin{framedexample}
    \englishexamples{}
    \begin{tabular}{l|l}
        %Hash:& 48d7a5 \\
        % English Translation &\\
        Before Previous Sentence: & We describe networks that do not use a shared transmission medium .\\
        Previous sentence: & Transmission speeds vary greatly depending on the type of network . \\
        Current Sentence Start: & They start at tens of kilobits per second , but also reach \\
        \textbf{Current Word:} & \textbf{speeds} \\
        Current Sentence End & of the order of a few gigabits per second . \\
        Next Sentence: & An example of such a network is the Internet.\\
        After Next Sentence: & Large networks include : \\
        Is Correct w.r.t. Cur. Sentence: & True \\
        Is Correct w.r.t. Whole Context: & False \\
        Contains Spelling Typo: & False 
    \end{tabular}
    \end{framedexample}
    
    \caption{Annotation item example. The annotator marks whether the word "rychlosti" is correct given a context of the current sentence, whether it is still correct in the context of two previous and two following sentences and whether it contains a typo.}
    \label{table:annotation_item}
\end{figure}

To create the annotation items, we concatenated data from all domains, both the original Wikipedia data \cite{naplava2018diacritics} and other domains (Section~\ref{sec:other_domains}) and we further considered those words in which the results of our system did not match target word. Before annotation, we automatically filtered out some cases:

\begin{itemize}
    \item Predictions, in which the system and the target words are variants (as marked by MorphoDita \cite{MorphoDiTa}) were automatically marked correct.
    \item Predictions, in which the target word was marked as non-existing by MorphoDiTa, while the system word was marked as Czech, were considered dubious and removed from our analysis.
\end{itemize}

For the remaining 4702 words, two annotation items were created: one with the predicted word and one with the gold reference word in the position of the annotated \textit{Current Word}. The annotation process took circa 70 hours.

The basic analysis of the annotated system errors is the following: There are 4702 wrongly diacritized words in the all our data concatenated. Annotations revealed that 960 of the mispredicted words contain a non-diacritical error and we do not consider them further, as mentioned above. The remaining 3742 mispredicted words can be categorized as follows:

\begin{itemize}
    \item System correct, Gold correct: 19\% (694 of 3742) -- plausible variants
    \item System correct, Gold wrong: 25\% (964 of 3742) -- system corrects data error
    \item System wrong, Gold wrong 1\% (31 of 3742) -- uncorrected error in data
    \item System wrong, Gold correct 55\% (2\,084 of 3742) -- real errors
\end{itemize}

\begin{table}[t]
    \centering
    \begin{tabular}{lccc}\toprule
    Domain  & Original & Annotated & Annotated w/o annotated typos \\\midrule
    Wiki & 99.22 & 99.49 & 99.66 \\ % 0.34615384615384054; 0.33
    Natives Formal & 99.50 & 99.75 & 99.75 \\ % 0.50; 0
    Natives Informal & 99.12 & 99.53 & 99.62 \\ % 0.46590909090908944; 
    Romi & 99.11 & 99.46 & 99.54 \\ % 0.3932584269662855
    Second Learners & 99.18 & 99.73 & 99.79 \\\bottomrule% 0.6707317073170753
    \end{tabular}
    \caption{Alpha-word accuracy of Czech model on 5 datasets from various domains.}
    \label{table:other_domain_results}
\end{table}

Interestingly, the annotations revealed that about 44\% of errors are not errors at all. In 694 cases (19\%) both the system word and the gold word are correct, which is justified by the plausible variants. In 964 cases (25\%) the original gold annotation was wrong whereas the system annotation was correct, which means that the system effectively corrected some of the errors in the original data. The remaining 31 cases are for neither the system nor the gold word being correct. Finally, the annotations confirmed 2084 real system errors, which we postpone for a more detailed analysis in the following Section~\ref{sec:real_errors}.

Plausible variants, which constitute 19\% of the annotated errors, are the most interesting item. Please note that our criterion for plausible variant was strict: only cases ambiguous both in the sentence and document context were marked as plausible variants. Circa 72\% percent of these words share a common lemma. As Table~\ref{table:basic_pos}.a and Table~\ref{table:extended_pos}.a show, singular/plural ambiguities by far most often arise in inanimate masc. genitive (\textit{programu/programů}, \textit{šachu/šachů}).
Another common ambiguity is passive participle vs. adjective (\textit{založena/založená}), generally known to be difficult for diacritization disambiguation \cite{NovyEncyklopedickySlovnik}. More interesting examples are given in Table~\ref{table:basic_pos}.a and Table~\ref{table:extended_pos}.a.

To conclude, we use the collected annotations to refine our previous results, which we display in Table~\ref{table:other_domain_results}. When considering all annotated words, including those preprocessed with MorphoDiTa, we achieve 35\% to 67\% error reduction. When omitting words newly marked by human annotators as containing another (non-diacritical) error, the error rate gets additionally reduced by up to 33\%.

\subsection{Analysis of Real Errors}
\label{sec:real_errors}

We follow with a morphological analysis of the remaining confirmed errors, which constitute 55\% of the annotated mispredictions. To determine the morphological categories of the erroneously predicted words, we use UDPipe~\citep{tsd2019_czech_nlp} to generate morphological annotations for all words in model hypotheses and gold sentences. We then inspect the most frequent confusions between the system and the gold morphological annotations of words, using the Universal POS tags and Universal features \cite{nivre-etal-2020-universal}.

% kolik je statistika pres pocet spatnych znaku
% 1 3317
% 2 390
% 3 29
% 4 5
% 9 1 (bug)

% - pokud formulujeme diakritizaci jako proposing oprav, tak je horsi dat spatnou opravu nez zadnou, zkusme se podivat, jak to  delame, tj. kolik dame spatnych a kolik dame zadnych

% bad 1510
% none 574

The annotations confirmed an interesting discourse phenomenon: a word can be correctly diacritized in multiple ways given the context of its sentence, however only a single correct diacritization variant exists if a wider context is taken into account. There are 50 such annotated cases; two examples are displayed in Figure~\ref{table:sentence_vs_whole_context_examples}. Although this phenomenon is interesting from a discourse perspective, its low proportion to actual errors (50 of 2084) indicates that it is quite rare. This implies that training models on longer texts (we currently train our model on examples comprising maximally 128 subwords -- see Section~\ref{ssec:training_details}) does not promise potential for overall improvement. Finally, we offer a categorization of such ambiguities by means of the Universal POS tags and Universal features \cite{nivre-etal-2020-universal} in Table~\ref{table:basic_pos}.b and Table~\ref{table:extended_pos}.b, respectively.

\begin{figure}[p]
    \footnotesize
    \begin{framedexample}
Tento motiv může být ovlivněn sibiřským šamanismem a průvodce pak má funkci psychopompa . 

Kromě bohů znali pohanští Slované i celou řadu \uline{nižších bytostí} , nazývány byly většinou slovem běs či div , které souvisí s indickým déva .

K nejvýznamnějším patří zmiňované \textbf{víly/vily} . 

V různých podáních existují \uline{víly} lesní , vzdušné , horské a také \uline{víly} zlé .  

Existují další ženské bytosti jim podobné , patří mezi ně především \uline{rusalky} , \uline{divé ženy} nebo \uline{divoženky} doprovázené \uline{divými muži} .

\leavevmode

Další \uline{dokumenty} týkající se Jana Žižky z Kalichu jsou \uline{dva listy} odeslané z kláštera ve Vilémově datované k 16. březnu a 1. dubnu 1423 .  

Slepý vojevůdce \uline{v nich} vyzývá své straníky z orebského svazu k poradě naplánované na 7. či 8. dubna\\do Německého Brodu . 

Z \textbf{dopisů/dopisu} je patrné , že se pokoušel dokonaleji zorganizovat husitskou vojenskou moc ,\\pro boj s domácím i zahraničním nepřítelem . 

O čtrnáct dní později Žižka spolu s orebity vedl válku se spojenci krále Zikmunda , zejména na Bydžovsku s panem Čeňkem z Vartenberka . 

Tohoto šlechtice s jeho leníky a spojenci porazil 20. nebo 23. dubna v bitvě u Hořic , načež dál pokračoval v plenění jeho zboží . 
\end{framedexample}

\itshape
\begin{framedexample}
\englishexamples{}

This motif can be influenced by Siberian shamanism , and the guide then has the function of a psychopomp . 

Apart from the gods, the pagan Slavs knew a number of \uline{lower beings} , mostly called Raver or Wonder , which is related to Indian deva .

Among the most important are the mentioned \textbf{fairies/villas}. 

There are wood \uline{fairies}, air \uline{fairies} , mountain \uline{fairies} , and also evil \uline{fairies} in various forms . 

There are other female beings similar to them , they include mainly \uline{mermaids} , \uline{wild women} or \uline{witches} accompanied by wild men .

\leavevmode

Other \uline{documents} concerning Jan Žižka of the Kalich are \uline{two letters} sent from the monastery in Vilémov dated March 16 and April 1 , 1423 .  

\uline{In them} , the blind military leader invites his party members from the Orebic Union to a meeting scheduled for April 7 or 8 in Německý Brod . 

The \textbf{letter shows/letters show} that he has tried to better organize Hussite military power , to fight both domestic and foreign enemies.

Fourteen days later , Žižka , together with the Orebits , waged war with King Zikmund's allies , especially in the Bydžov region with Mr. Čeněk of Vartenberk .

He defeated this nobleman with his feoffees and allies on April 20 or 23 at the Battle of Hořice , after which he continued to plunder his goods . 
\end{framedexample}

    \caption{Two examples of ambiguous diacritization determined by document context.}
    \label{table:sentence_vs_whole_context_examples}
\end{figure}

\newcommand{\subcaption}[2][1.5em]{\vspace{.2em}\\\multicolumn{3}{c}{#2}\vspace{#1}}

\begin{table}[p]
\centering
\begin{tabular}{lrl}\toprule
\textbf{Type} & \kern-.5em\textbf{Count}\kern-.45em & \textbf{Examples} \\\midrule

NOUN $\leftrightarrow$ NOUN & 406 & program[uů], šach[uů], text[uů] \\
ADJ $\leftrightarrow$ ADJ & 162 & znám[áa], založen[aá], schopn[ií] \\
ADV $\leftrightarrow$ ADJ & 59 & stejn[ěé], krásn[ěé], běžn[ěé] \\
PROPN $\leftrightarrow$ PROPN & 31 & Aristotel[eé]s, Sokrates/Sókratés, J[aá]n \\
VERB $\leftrightarrow$ VERB & 20 & zamýšlím/zamyslím, odráží/odrazí, os[ií]dlují \\
ADJ $\leftrightarrow$ VERB & 3 & vznikl[áa], rádi/radí, splaskl[áa] \\
NOUN $\leftrightarrow$ ADJ & 2 & přesvědčen[íi], očištěn[íi] \\
ADJ $\leftrightarrow$ NOUN & 2 & veden[ií], považován[ií] \\
DET $\leftrightarrow$ DET & 2 & jej[íi]ch, svoj[íi]
\subcaption{(a) Plausible variants.}\\

\textbf{Type} & \kern-.5em\textbf{Count}\kern-.45em & \textbf{Examples} \\\midrule
NOUN $\rightarrow$ NOUN & 32 & stát/stať, objekt[uů], pulsar[uů] \\
VERB $\rightarrow$ VERB & 4 & narazí/naráží, řekn[ěe]te, žij[íi] \\
DET $\rightarrow$ DET & 3 & jej[ií]ch \\
ADJ $\rightarrow$ ADV & 3 & současn[éě], pravé/právě, praktick[ýy] \\
ADJ $\rightarrow$ ADJ & 2 & znám[áa], žádanou/zadanou \\
ADV $\rightarrow$ ADJ & 2 & stejn[ě/é] \\
NOUN $\rightarrow$ VERB & 1 & mysl[ií]
\subcaption{(b) Disambiguation from document context.}\\

\textbf{Type} & \kern-.5em\textbf{Count}\kern-.45em & \textbf{Examples} \\\midrule
NOUN $\rightarrow$ NOUN & 1596 & stát/stať, lid[íi], program[uů] \\
PROPN $\rightarrow$ PROPN & 587 & Aristotel[eé]s, Sokrates/Sókratés, Kast[ií]lie \\
ADJ $\rightarrow$ ADJ & 521 & znám[aá], založen[aá], říd[ií]cí \\
VERB $\rightarrow$ VERB & 193 & m[ůu]že, M[aá]m, m[aá] \\
ADJ $\rightarrow$ ADV & 134 & krásn[éě], hezk[ýy], dobré/dobře \\
PRON $\rightarrow$ PRON & 129 & j[íi], n[íi], n[íi]ž \\
ADV $\rightarrow$ ADJ & 112 & stejn[ěé], pěkn[ěé], Obvykl[eé] \\
DET $\rightarrow$ DET & 59 & jej[íi]ch, svoj[ií], naš[ií] \\
NOUN $\rightarrow$ ADJ & 47 & mobiln[ií], brány/braný, češka/česká
\subcaption[0pt]{(c) Real errors.}\\\bottomrule
\end{tabular}
    \caption{Error categorization with universal POS. The context-dependent morphological annotations were obtained automatically using UDPipe.}
    \label{table:basic_pos}
\end{table}

\begin{table}[p]
\centering
\small
\begin{tabular}{lrl}\toprule
\textbf{Type} & \kern-.5em\textbf{Count}\kern-.45em & \textbf{Examples} \\\midrule
Number & 325 & program[uů], šach[uů], objekt[uů] \\
\makecell[l]{Passive participle / adjective\\~~+ more features} & 116 & založen[aá], vzdálen[aá], nazývan[aá]  \\
Lemma & 82 & l[eé]ty, mas[ií]vu, p[ée]rových \\
Adj $\leftrightarrow$ Adv  & 59 & stejn[éě], krásn[éě] \\
Variant + more features & 31 & znám[áa], schopn[ií], spokojen[íi] \\
Case & 25 & dr[aá]hami, dr[aá]hách, č[aá]rou \\
Lemma + more features & 21 & zamýšlím/zamyslím, ná[sš], pacht[uů]  \\ % 92 - 82 + 11
Lemma, NameType & 20 & Aristotel[eé]s, Sokrates/Sókratés, [ÍI]lias \\
Case, Number & 8 & boh[ůu], násobk[uů], funkc[íi] \\
Number, Person & 5 & považuj[íi], věnuj[ií], kupuj[ií]
\subcaption{(a) Plausible variants.}\\

\textbf{Type} & \kern-.5em\textbf{Count}\kern-.45em & \textbf{Examples} \\\midrule
Lemma + more features & 15  & stát/stať,  tvář/tvar, pravé/právě \\
Number & 15 & objekt[ůu], pulsar[uů], muzikál[ůu] \\
Lemma & 6 & řazení/ražení, v[ií]ly \\
Adj $\leftrightarrow$ Adv & 4 & stejn[ěé], současn[éě], praktick[ýy] \\
Case, Gender, Number & 3 & jej[ií]ch \\
Number, Person & 2 & narazí/naráží
\subcaption{(b) Disambiguation from document context.}\\

\textbf{Type} & \kern-.5em\textbf{Count}\kern-.45em & \textbf{Examples} \\\midrule
Lemma + more features & 924  & stát/stať, [čc], [žz]e \\ % 879 + 25 + 20
\makecell[l]{Lemma, named entity\\~~+ more features} & 382 & D[ií]ogenés, Hal/\Hwithstroke{}al, Dvořák/Dvorak \\
Number & 226 & milion[uů], reproduktor[ůu], dokument[ůu] \\
Case & 149 & j[ií], n[íi], zem[íi] \\
Adj $\leftrightarrow$ Adv & 132 & pěkn[éě], česk[ýy], současn[éě] \\
\makecell[l]{Passive participle / adjective\\~~+ more features} & 37 & spojen[aá], pojmenovan[áa], prodaný/prodány \\
Case, Number & 27 & referent[uů], Dvořák[ůu], akademi[íi] \\
Case, Gender, Number & 16 & jej[íi]ch, j[íi]m \\
Number, Person & 15 & píš[ií], pracuj[ií], žij[íi] \\
Variant + more features & 8 & znám[áa], schopn[áa], hodn[áa]
\subcaption[0pt]{(c) Real errors.}\\\bottomrule
\end{tabular}
    \caption{Error categorization with extended Universal Features. The first column (Type) is the (primary) difference between the context-dependent feature sets of the system word and the gold word.}
    \label{table:extended_pos}
\end{table}

% Stát Stať
% 2 ) Narození Krista         
% 2 ) Srážka s autem
% 2 ) Stať : a ) Představení
% 2 ) Stať : a ) Vnější charakteristika
% 2 ) Stať : a ) Člověk a příroda

% Mysli Myslí
% *Mysleli* že jestli ě člověk nemocný , to není možné nic udělat jenom modlit se . Myslí , že kdy někdo nosí oblek , je šikovný .
% Myslí na Slovensko .
% Myslí na další část dne a planujou kino .
% Myslí na dobré počasy a hezký den .

% vily víly                                                                         
% Tento motiv může být ovlivněn sibiřským šamanismem a průvodce pak má funkci psychopompa .
% Kromě bohů znali pohanští Slované i celou řadu nižších bytostí , nazývány byly většinou slovem běs či div , které souvisí s indickým déva .
% K nejvýznamnějším patří zmiňované víly .
% V různých podáních existují víly lesní , vzdušné , horské a také víly zlé .
% Existují další ženské bytosti jim podobné , patří mezi ně především rusalky , divé ženy nebo divoženky doprovázené divými muži .

The remaining errors are a mix of complicated disambiguation cases or rare named entities. The most frequent errors bear similarity to plausible variants (compare Table~\ref{table:extended_pos}.a and Table~\ref{table:extended_pos}.c), only with a different order of appearance.  Unlike plausible variants (Table~\ref{table:extended_pos}.a), most frequent mismatches occur already at the level of lemmas (\textit{stát/stať}, \textit{že/ze}, see Table~\ref{table:extended_pos}.c). Second most frequent cases are rare named entities (\textit{Sokrates/Sókratés}, \textit{Aristoteles/Aristotelés}, \textit{Diogenés/Díogenés}). Number is again often hard to disambiguate in inanimate masc. genitive (\textit{milionu/milionů}, \textit{reproduktoru/reproduktorů}, \textit{dokumentu/dokumentů}), followed by fem. case (\textit{ji/jí}, \textit{ni/ní}, \textit{zemi/zemí}).

\section{Conclusion}

We implemented a model for diacritics restoration based on BERT that outperforms previous state-of-the-art models. Further analysis on Czech data collected from additional, noisy domains shown that the model exhibits strong performance regardless the domain of the data. 

We further annotated all reported mispredictions in Czech and found out that more than one correct variant is sometimes possible. Rarely, disambiguation on document level is necessary to distinguish between variants correct within the sentence context. We elaborated on these phenomena using morphological annotations and utilized them to further analyse real confirmed errors of the systems. 

As for future work, we propose experimenting with a single joint model for a subset of languages, despite our initial unsuccessful attempts at training a single model for all languages, including an introduction of a larger XLM-Roberta model~\cite{conneau2020unsupervised}.

\section*{Acknowledgements}

The work described herein has been supported by and has been using language resources  stored by LINDAT/CLARIAH-CZ project of the Ministry of Education, Youth and Sports of the Czech Republic (project No. LM2018101) and also by OP VVV LINDAT/CLARIAH-CZ EXTENSION project of the Ministry of Education, Youth and Sports of the Czech Republic (project No. \hbox{CZ.02.1.01/0.0/0.0/18\_046/0015782}). It has further been supported by the Grant Agency of the Czech Republic, project \hbox{EXPRO} LUSyD (GX20-16819X), and Mellon Foundation project No. G-1901-06505, by SVV project number 260 575 and by GAUK 578218 of the Charles University.

We are very grateful to our anonymous reviewers for their valuable comments and corrections.

% Bibliography
% ============
% You may either enter the bibliography manually, possibly making use of
% \label and \ref, or use BibTeX. The bibliography style is set
% automatically. You process the bibliography by BibTeX in the
% standard way and include it by:
\bibliography{mybib}

% If needed, add appendices here
%\section*{Appendix A: \ldots}

\correspondingaddress
\end{document}